# Detection of Cyberbullying in GIF using AI

Pal Dave and Xiaohong Yuan and Madhuri Siddula and Kaushik Roy

North Carolina A&T State University, Greensboro NC 27405, USA
pvdave@aggies.ncat.edu

**Abstract.** Cyberbullying is a well-known social issue, and it is escalating day by day. Due to the vigorous development of the internet, social media provide many different ways for the user to express their opinions and exchange information. Cyberbullying occurs on social media using text messages, comments, sharing images and GIFs or stickers, and audio and video. Much research has been done to detect cyberbullying on textual data; some are available for images. Very few studies are available to detect cyberbullying on GIFs/stickers. We collect a GIF dataset from Twitter and Applied a deep learning model to detect cyberbullying from the dataset. Firstly, we extracted hashtags related to cyberbullying using Twitter. We used these hashtags to download GIF file using publicly available API GIPHY. We collected over 4100 GIFs including cyberbullying and non-cyberbullying. we applied deep learning pre-trained model VGG16 for the detection of the cyberbullying. The deep learning model achieved the accuracy of 97%. Our work provides the GIF dataset for researchers working in this area.

## 1 Introduction

Traditional bullying occurs in person or face to face or within physical environments such as schools, workplaces or neighborhoods. Bullying refers to the act of intentionally causing harm, distress or discomforts to another person through physical or verbal means and even relational. Physical bullying involves aggression such as hitting, pushing, kicking to another person and even damaging their belongings. Verbal bullying includes teasing, provoking, insulting, calling names and using inappropriate language to belittle the victim. Relational bullying is spreading rumors and manipulating someone by the talks. Cyberbullying is related to technology and its digital nature. Cyberbullying takes place through electronic devices and social media platforms. Cyberbullying allows perpetrators to remain anonymous or use a fake identity, making it easier for them to target victims. It impacts victims mentally and emotionally. The technology provides a sense of detachment to perpetrators which leads them to behave in a way they would not in offline situations. Research has shown that intentionally or unintentionally most of the adolescents are more involved in activities which are associated with cyberbullying on social mediums, due to the lack of knowledge about their actions and understanding of how it can leave an impact to others.

Social media have become a popular platform for people to express and share their opinion. Many social media users are negligent and neglectful in considering ethics and social norms. As a result, cyberbullying has become a serious issue for every age group. Research has shown a spike in cyberbullying cases in recent years and it has been predicted that cyberbullying will keep growing in the near future. The actual cyberbullying incidents can be higher than what research shows since some victims do not report the incident or try to hide it. Gassem et al. [10] conducted a study on 335 students, aged between 12-18 years using validated online questionnaire. Based on their study they concluded 20% or more spend 12 hours online on daily bases and 44.7% have experienced cyberbullying. Sunawan et al. [11] presented a study of 103 respondents aged around 19-24, who were indicated as perpetrator of social media. They collected data using emotion regulation questionnaire and the cyberbullying inventory. They found that, the correlation between cyber victimization and cyberbullying will strengthen if there is weak level of emotion regulation. In a survey by Mallory N. et al. [12] on cyberbullying, 1 in 5 parents worldwide say their child has experienced cyberbullying; 1 in 3 parents reported about it.

It is important to detect cyberbullying on different kinds of data such as Text, Images and GIFs/Stickers. Artificial Intelligence and machine learning have been used for detecting cyberbullying in text-based data [5],[4],[7]. Vishwamitra et al. [8] proposed to use feature identification and extraction, and convolutional neural network model (CNN) to detect cyberbullying in images. To detect text-based cyberbullying, Luo et al. [4] developed a framework GCA model which is based on bidirectional gated recurrent unit, attention layer and convolutional neural network. Guo et al. [9] proposed a vision and language multi-model to detect hateful meme over the social media on a covid-19 case study. They showed that textual modality provides valuable information to detect new types of hateful meme.

Most existing research uses textual and image data for cyberbullying detection. GIFs/Stickers are gaining popularity on social medium. A GIF is known as Graphics. There are three types of GIFs: 1) Clips/video; 2) Animation of 3-4 seconds; 3) Stickers. GIFs can be created by anyone with any content since some social media allows users to make a GIF if the video, they want to share is small in size. A popular website for GIFs and stickers is GIPHY. It is predicted that GIPHY serves 7 billion GIFs and stickers to 500 million people every single day. According to a survey conducted by Baretree Media [14], 71% Americans would rather send a digital sticker over a block of text. Miltner et al. [13] state that GIFs are highly versatile and they have become a key communication tool in digital culture due to their features. Some GIFs include textual data, and some GIFs include facial expression and textual data. However, there is little research on detecting cyberbullying on GIF/Stickers. In this research we have undertaken the integration of GIFs/stickers into our methodology. We developed a new dataset of GIFs for our study, We employed the pre-trained VGG16 for the cyberbullying detection. This research considers the evolving nature of online communication and the significance of multimedia elements in cyberbullying detection. Our research focus is to enhance the accuracy and effectiveness of cyberbullying detection in online visual content.

The rest of the paper is organized as follows: Section 2 describes the related work on cyberbullying detection. Section 3 introduces the proposed VGG16 model to classify cyberbullying. Section 4 reports our experiments and results, Section 5 gives conclusion and discusses future work and section 6 is acknowledgement.

## 2     Related Work

Our work relates to cyberbullying detection on textual data and image data, and the use of deep learning models for cyberbullying detection. In this section, we review related work in these areas.

Cyberbullying can be done through all means of information exchange using social media. One of the common means of cyberbullying is using textual data such as direct messages, comments and tweets.

Alholoul et al. [1] proposed a frame which is a combination of attention layer and convolutional layer to extract cyberbullying related keywords from tweets and categorize them into different classes such as age, ethnicity, gender and more. Perera et al. [3] presented an automatic system to detect cyberbullying. The architecture relies on cyberbullying text along with themes/categories associated with cyberbullying such as racist, physical mean, swear etc. They focused on traditional feature extraction methods which helped them achieve higher accuracy using support vector machines and logistic regression.

Cyberbullying appears differently in different social media platforms. Dadvar et al. [6] approached this issue by looking at three different social media datasets. They used the existing dataset of Wikipedia, Twitter and Form spring. They also developed a new dataset which is from YouTube but it also includes textual data. They used word embeddings and deep learning models to detect cyberbullying.

Images are one of the ways cyberbullying can occur. Vishwamitra et al. [8] proposed a framework where they extracted features in the form of gestures, body pose, facial emotions and then they used multimodal model to get the results. They leveraged the VGG16 pre-trained model for feature extraction, the model combines low-level image features with identified factors to achieve precise classification of cyberbullying and non-cyberbullying images. The VGG16 model produces 512 convolutional feature maps, which are fused with the output of multi-layer perceptron model using late fusion. The combined feature vectors from both models are jointly utilized to classify images. In contrast to the preceding research, which focuses on the classification of static images into cyberbullying and non-cyberbullying, our technique concentrates on GIFs to classify cyberbullying and non-cyberbullying.

Shirzad et al. [15] proposes a novel multimodal sentiment analysis system that integrates text, image, and GIF modalities, specifically designed for tweets. Using the Twitter search API, the VADER tool is used to preprocess text data for sentiment analysis, and the VGG16 network is optimised for sentiment analysis of images. Additionally, a novel approach to GIF sentiment analysis is presented, which combines facial emotion recognition with visual sentiment analysis within GIF frames. The outcomes demonstrate how well the framework performs sentiment analysis across a range of modalities, highlighting its thorough and creative approach to social media sentiment analysis. Unlike the previous framework mentioned, our method focuses solely on GIFs to detect cyberbullying, extending the application of multimodal sentiment analysis in the context of text, image, and GIF modalities. While both studies involve dividing GIFs into frames and use VGG16 for analyzing GIFs, our work differs by focusing specifically on detecting cyberbullying content within GIFs.

Mekala et al. [2] shared their views about cyberbullying by their own experience, they said it lowered self-esteem and it escalated their suicidal thoughts. They used twitter's tweets as dataset to train the model using BERT (Bidirectional Encoder Representations from Transformers) model and then classified the tweets to detect sentiments.

Most of the cyberbullying detection research has been focused on textual and image data. With the recent popularity of animated GIFs/stickers, we focused on building a new dataset for cyberbullying detection on GIFs.

## 3 Proposed Methods

In this section, we introduced the work done for collecting the GIF data set. We introduce the data collection, data cleaning and data annotation scheme. The proposed model for cyberbullying detection is also describe.

### 3.1 Dataset Collection

From our literature review we realize there are not many GIF datasets available for detecting cyberbullying. There are some datasets used for sentiment analysis not for cyberbullying. We require a dataset in order to conduct research on detecting cyberbullying from GIFs. To collect the GIF dataset, we first collected hashtags. From twitter, we manually searched for trending tweets which are related to cyberbullying and the hashtags used within. Initially we started with 61 cyberbullying-related hashtags, which we found in cyberbullying related tweets. We used GIPHY API to search for GIFs using the hashtags we collected from Twitter.

GIPHY platform is widely integrated to many popular social media platforms. By using GIPHY, GIFs are easily accessible for users to search, share and use animated GIFs and stickers. Some of the major social media platforms have integrated GIPHY's services include: Facebook and Meta, Twitter, WhatsApp, Instagram, Snapchat and TikTok. These social media allow users to use GIFs from GIPHY API to interact on social media in the form of comments, direct messages, and stories. GIPHY API can be used to download GIFs by using hashtags. Fig. 1 shows the examples of hashtags, which are as follows: Cyberbullying: bullying, hateLGBTQ, whitetrash, racist and bullyingiscool; Non-cyberbullying: Awesome, good work, how are you doing? We collected cyberbullying related GIF using hashtags extracted from tweets. For non-cyberbullying related GIF, we used conversational hashtags which do not represent bullying.

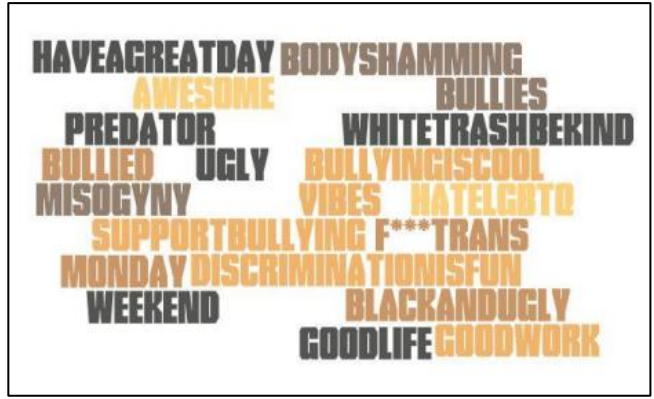

**Fig. 1.** Word cloud of extracted hashtags

Fig. 1 showcase a word cloud visualizing extracted hashtags related to cyberbullying and non-cyberbulling, generated using Powerpoint presentation software. It provides a graphical representation of extracted hashtags dataset. We restricted our dataset to English language. All the collected GIFs which include textual data has text in English language. We also checked for duplicates in the dataset. In total we have 4,100 GIFs including cyberbullying and non-cyberbullying. Table 1 delineates the distribution of GIFs across cyberbullying and non-cyberbullying classes.

**Table 1.** Summary of Dataset

|  | GIFs |
| --- | --- |
| Cyberbullying | 1669 |
| Non-Cyberbullying | 2431 |
| **Total** | **4100** |

### 3.2 Annotation Scheme

After collecting the dataset, we annotated the dataset as cyberbullying or non-cyberbullying. In our annotation scheme, we annotated a GIF as cyberbullying if it has one of the following: (1) Bullying directed towards an individual, race or organization 18. (2) Hate speech towards victims or unethical behavior and comments on personal appearance which includes cursing and inappropriate remarks. (3) Gestures and expression of the individual which includes hand gestures, expressions of repugnance, hostility. We annotated GIFs in two steps. In the first round, four high school students independently labelled a set of 250 GIFs each. They had to classify the GIFs into two categories: cyberbullying and non-cyberbullying. We also annotated the same 1,000 GIFs independently. In the second round, we checked the results of both annotations which were nearly identical. After the second round, we did our final annotations of the remaining 3,100 GIFs.

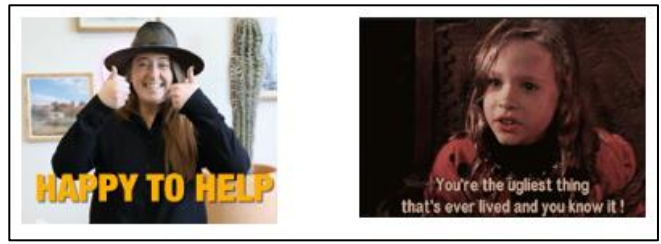

**Fig. 2.** Examples of Cyberbullying and Non-Cyberbullying GIFs from collected dataset

**Data Pre-processing.** After annotation of the dataset, we pre-processed the GIF dataset into frames. We used the Python library OpenCV and PIL (Python Imaging Library) to convert GIF URLs into a set of up to 16 frames. The GIFs have different sizes. Therefore, there are a different numbers of frames per GIF. For example, the highest number of frames we have is 236, and the lowest is 2.

*Removing duplicate Frames.* In the set of frames for each GIF, we noticed two or more corresponding copies of the same structures in a dataset. And some frames have discrepancies in textual data. We decided to remove blurred content or unclear textual data, resulting in a new dataset. Our proposed model is trained on this dataset. To exclude unnecessary frames, we choose to extract GIF frames based on the size of the GIF, and the highest number of frames is 16 for every GIF which exceeds the length. By applying this, we were able to reduce the number of frames.

*Cleaning the Dataset.* Cleaning the dataset helps with reducing the inconsistencies and inaccuracies in the data. A cleaned dataset contributes to better model performance. The dataset includes three types of data: (1) GIFs which have only textual data (2) GIFs which have no textual data (3) GIFs which have both human facial expressions and textual data We used GIFs which have both human facial expressions and textual data to train our deep learning model. After cleaning the dataset, we have a total of 4,220 frames. Table 2 provides an overview of the dataset's composition post cleaning.

**Table 2.** Summary of Dataset after Cleaning

| Dataset | After Cleaning |
| --- | --- |
| Cyberbullying | 1789 (Frames) |
| Non-Cyberbullying | 2431 (Frames) |

*Cleaning of the Dataset.* After cleaning the dataset, we applied a series of techniques to augment our dataset. It includes ImageDataGenrator class, rotation range, width shift range, height shift range, shear range, zoom range, horizontal flip and fill mode to increase the size of the training dataset by applying various transformations to the images. After applying data augmentation techniques, we have a total of 16,875 frames to train our dataset. We opted for a specific category of GIFs (involving human facial expressions and textual data) in our training process, anticipating improved outcomes as a result of using similar GIF types. We trained our dataset using deep learning pre-trained VGG16 model on 16,875 frames.

*Proposed Model.* Fig.3 shows the proposed system architecture for cyberbullying detection. To gather relevant data, we leveraged Twitter hashtags associated with cyberbullying and utilized the GIPHY API to acquire cyberbullying and non-cyberbullying related GIFs. Subsequently, the collected GIFs were pre-processed, extracting frames up to the 16th frames for each GIF. After that we employed data augmentation techniques to expand the dataset, ensuring robust model training. To evaluate our model, we used our meticulously collected dataset, we used the dataset for our training, validation and testing by splitting this data into 0.8, 0.1, 0.1 ratio. In our system architecture the pre-trained VGG16 deep learning model served as the foundation. This model underwent a process known as fine-tuning, where it was adapted to proficiently classify GIFs into categories of cyberbullying and non-cyberbullying. The utilization of VGG16 in this approach is to achieve accurate identification of cyberbullying instances. The pre-trained VGG16 model brings knowledge and features learned from a diverse range of images, which proves beneficial in enhancing the model's ability to categorize visual content present in GIFs. The VGG16 model has been proven to demonstrate strong performance in image processing. [8] [15]

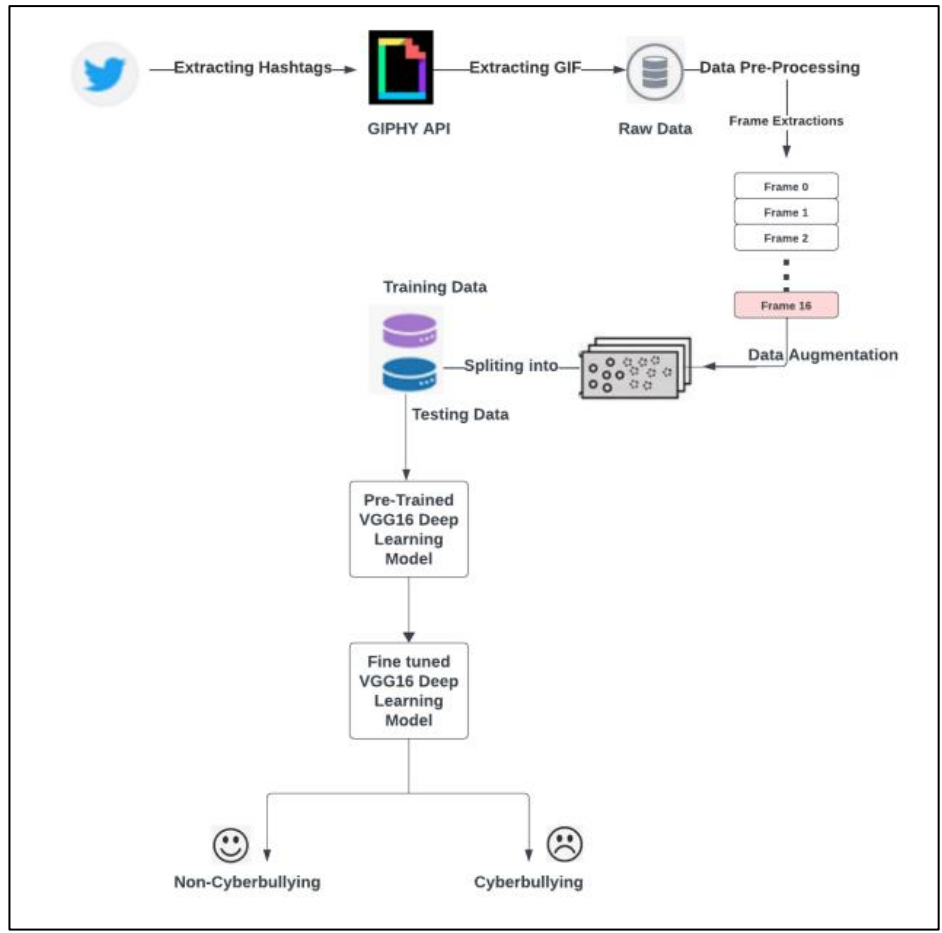

**Fig. 3.** System Architecture

Fig. 4 shows our model architecture, which is based on a pre-trained VGG16 deep learning model. In this experiment, we aim to understand the performance of the VGG16 model in cyberbullying detection from GIFs.

We present a fine-tuned pre-trained VGG16 model for cyberbullying and non-cyberbullying GIF classification. Our approach is to use transfer learning to adapt the capabilities of the VGG16 model, such as feature extraction to detect cyberbullying from GIFs.

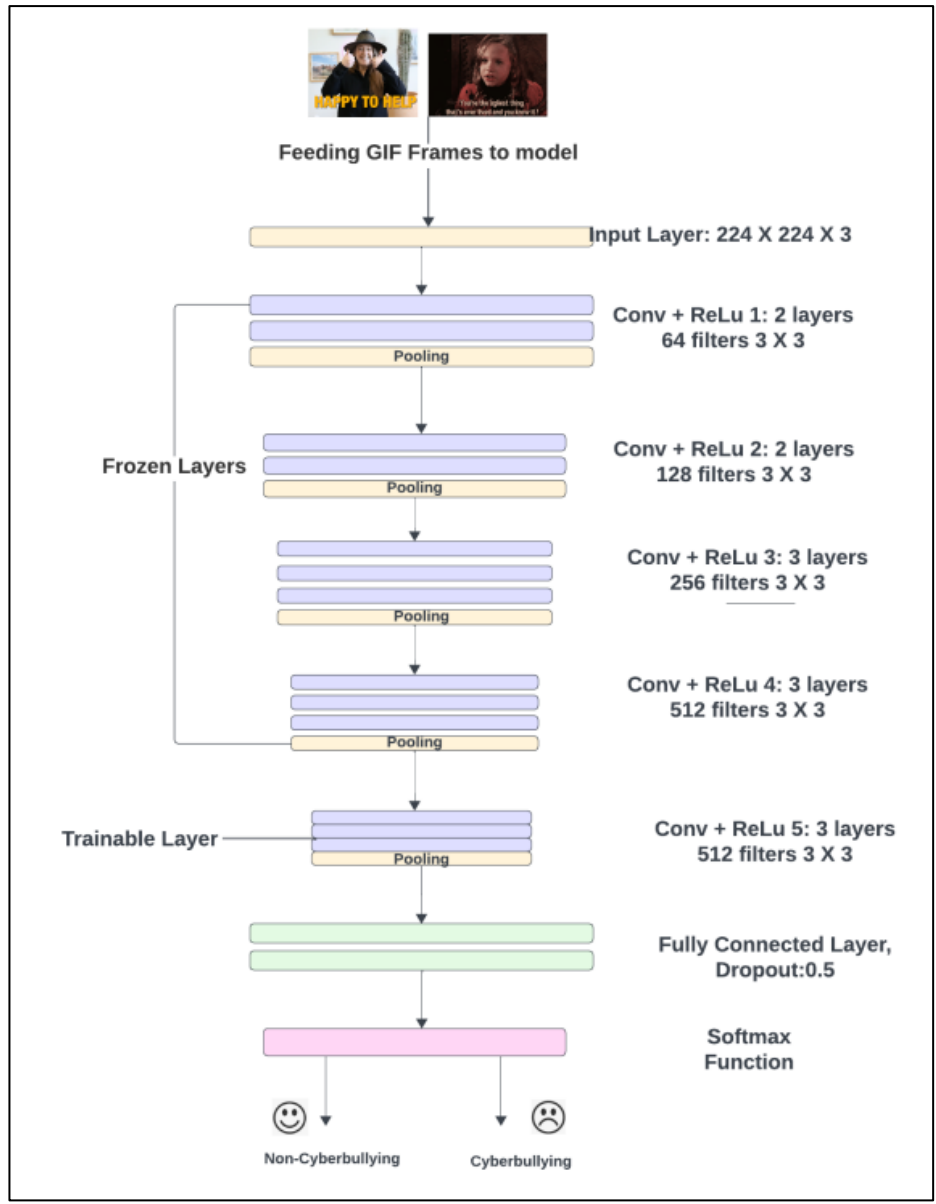

**Fig. 4.** Detailed Architecture of VGG16 Model

In the VGG16 model architecture, we initialized our base model on the ImageNet dataset with pre-trained weights. We removed the top 15 classification layers of the VGG16 model as we replaced them with our targeted classification needs. We applied a global average pooling layer to the output, followed by a fully connected layer of 256 units, employing the ReLU (Rectified linear unit) activation function. The last layer consists of a SoftMax activation and the number of distinct classes in our dataset. The replaced layers can be summarized as follows: 1 Global Average Pooling Layer: This layer reduces the dimension of the dataset and captures important features compactly. 2 Dense Layer with ReLU Activation: This layer has 256 units. This layer is responsible for feature extraction and learns higher-level features from the pooled features of the dataset. 3 Output Layer with SoftMax Activation: The last layer consists of the classes represented in the dataset. This SoftMax activation function indicates probabilities of classes and allows the model to make predictions.

## 4    Experiments and Results

We present analysis of our fine-tuned VGG16 model and cross validation on classification of cyberbullying in GIFs. The evaluation includes accuracy, confusion matrix, accuracy and loss curve. it helps to provide insights of our proposed model.

Table 3. Results of Cyberbullying Detection on GIFs using pre-trained VGG16 Model

| **Accuracy** | **0.9662** | | | |
|---|---|---|---|---|
| Class | Precision | Recall | F1-score | Support |
| Cyberbullying | 0.9589 | 0.9821 | 0.9704 | 951 |
| Non-cyberbullying | 0.9762 | 0.9457 | 0.9607 | 737 |
| Macro Avg | 0.9676 | 0.9639 | 0.9656 | 1688 |
| Weighted Avg | 0.9665 | 0.9662 | 0.9662 | 1688 |

      Table 3 shows the outcomes of our cyberbullying detection model applied to GIFs using the pre-trained VGG16 model. The accuracy metric 96.62% implies the overall correctness of classification of cyberbully and non-cyberbully instances. For cyberbullying class 95.89% (precision) out of the instances predicted as cyberbullying were accurate. Recall 98.21% states the percentage of correctly identified actual cyberbullying instances. F1 score 97.04% shows the in overall balanced performance for the cyberbullying class. For non-cyberbullying class, it showed similar results as cyberbullying class. Macro average metrics (96.76%,96.39%, 96.62%) represent the average performance across both classes. The weighted average metrics (96.65%,96.62%, 96.62%) consider the class imbalance, provides performance measure of distribution of instances across bullying and non-bullying classes.

      After the initial training of the VGG16 model, we employed cross-validation to assess the reliability of its performance. We implemented k-fold validation for training a VGG16 model on the dataset. The training dataset go through K-fold cross validation (k=5), where the model is trained on k-1 folds and validated on the remaining fold in each iteration. For each fold, a VGG16 model is trained on a subset of the training data for 50 epochs. We used early stopping method for hyper-parameter tuning.

In this case, the model will save weights at each epoch if the training loss improves. We used a ReduceLROnPlateau callback function to adjust learning rate and to avoid model over-fitting. To address imbalanced class distribution in training dataset, we included class weights to eliminate model biases. We used a transfer learning approach using VGG16, with strategies for distributed training, regularization and optimization of the model.

**Table 4.** Results of Cyberbullying Detection on GIFs using K-Fold Cross Validation Method

| Accuracy | 0.97 | | | |
|---|---|---|---|---|
| Class | Precision | Recall | F1-score | Support |
| Cyberbullying | 0.96 | 0.98 | 0.97 | 2050 |
| Non-cyberbullying | 0.98 | 0.96 | 0.97 | 2078 |
| Macro Avg | 0.97 | 0.97 | 0.97 | 4128 |
| Weighted Avg | 0.97 | 0.97 | 0.97 | 4128 |

Table 4 is the outcome of 5-fold cross validation technique. it shows higher level of accuracy and balanced performance in discriminating between cyberbully and non-bully instances in GIFs as indicted by precision, recall and F1-score metrics for each level. The use of 5-fold cross validation validates the model's generalizability and effectiveness in cyberbullying detection across dataset. We can see results in Table 4 are similar to those Table 3, which verifies the quality and stability of our model.

**Table 5.** Confusion Matrix of pre-trained VGG16 Model

| | Bullying | Non-bullying |
|---|---|---|
| Cyberbullying | 934 | 17 |
| Non-cyberbullying | 40 | 697 |

Table 5 presents the confusion matrix of the VGG16 model. It offers insights into the actual and predicted classifications for both bullying and non-bullying instances. The model correctly predicted 934 true positive instances as cyberbullying. There are 17 false positive instances predicted for bullying, but the instances were actually non-bullying. There are 697 true negatives instances predicted as non-bullying. And 40 false negative instances predicted non-bullying, but the instances were actually bullying.

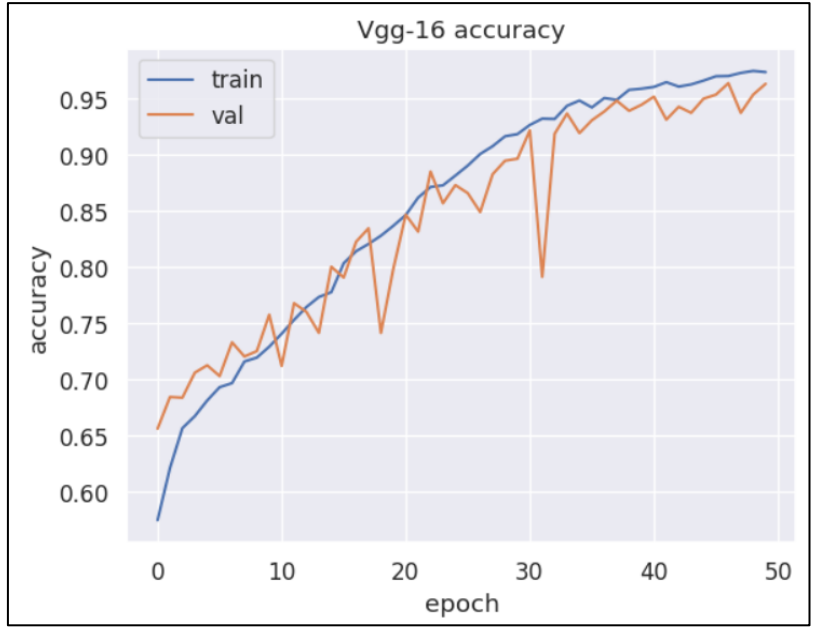

**Fig. 5.** Accuracy curve of Training and validation of pre-trained VGG16 Model

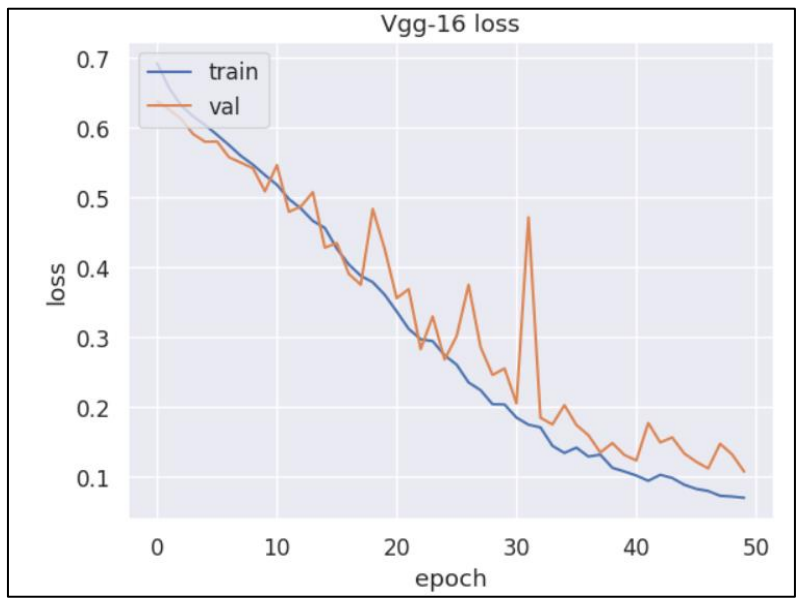

**Fig. 6.** Loss curve of Training and validation of pre-trained VGG16 Model

The fig. 5 and fig. 6 depict the dynamic progression of accuracy and loss curves during training and validation phases of the pre-trained VGG16 model, providing overview of its learning dynamics.

## 5 Conclusion

The dynamic and volatile nature of GIFs poses a significant challenge in the accurate detection of cyberbullying. The same GIF files can be interpreted differently by individuals, leading to diverse meanings and perspectives. In this paper, we described our work to detect cyberbullying in GIFs using a pre-trained VGG16 deep learning model. We manually collected our data using GIPHY API and annotated the dataset. We used GIFs which have both human facial expressions and textual data to train our deep learning model. Based on the results we analyzed, the precision and recall are high for both the bully and non-bully classes, which indicates the model can effectively identify cyberbullying instances from GIFs. In future work, we plan to collect more data and extract textual data stored in GIFs. By including textual data from the GIFs higher accuracy for the classification of cyberbullying and non-cyberbullying can be achieved.

## 6 Acknowledgement

This work is supported in part by the National Science Foundation (NSF). Any opinions, findings, and conclusions or recommendations expressed in this material are those of the author(s) and do not necessarily reflect the views of NSF.